\documentclass[runningheads]{llncs}

\usepackage{graphicx}
\usepackage{amssymb}
\usepackage{amsmath}
\usepackage{multirow}
\usepackage{algorithm}
\usepackage{algorithmic}
\usepackage{subfigure}
\usepackage{booktabs}
\usepackage{bbding}
\usepackage{threeparttable}
\usepackage{hyperref}
\usepackage{courier}

\author{Anxun He\orcidID{0000-0003-4848-0325} \and
Jianzong Wang\orcidID{0000-0002-9237-4231}\inst{(}\Envelope\inst{)}\and
Zhangcheng Huang\orcidID{0000-0001-6563-7668}\and
Jing Xiao}

\authorrunning{A. He et al.}
 
\institute{\normalsize Ping An Technology (Shenzhen) Co., Ltd., Shenzhen 518000, P.R China\\
\Envelope ~Corresponding Author: Jianzong Wang, \email{jzwang@188.com}}

\begin{document}
\title{FedSmart: An Auto Updating Federated Learning Optimization Mechanism}
\maketitle % typeset the header of the contribution
\begin{abstract}
Federated learning has made an important contribution to data privacy-preserving. Many previous works are based on the assumption that the data are independently identically distributed (IID). As a result, the model performance on non-identically independently distributed (non-IID) data is beyond expectation, which is the concrete situation. Some existing methods of ensuring the model robustness on non-IID data, like the data-sharing strategy or pre-training, may lead to privacy leaking. In addition, there exist some participants who try to poison the model with low-quality data. In this paper, a performance-based parameter return method for optimization is introduced, we term it \texttt{FederatedSmart} (\texttt{FedSmart}). It optimizes different model for each client through sharing global gradients, and it extracts the data from each client as a local validation set, and the accuracy that model achieves in round $t$ determines the weights of the next round. The experiment results show that \texttt{FedSmart} enables the participants to allocate a greater weight to the ones with similar data distribution.

\end{abstract}

\section{Introduction}

Securing high-quality machine learning models while working with different data owners is a challenge with user data security and confidentiality~\cite{yang2019federated}. In the past, there have been many attempts to address user privacy issues when exchanging data. For example, Apple recommends using Differential Privacy (DP) to respond these concerns~\cite{green2016apple}. The basic idea is to add appropriately calibrated noise to data in order to eliminate the identity of any individual but still retain the statistical characteristics~\cite{dwork2006differential}. However, DP can only prevent user information leakage to a certain extent. In addition, it is lossy in machine learning framework because the model built with noise is injected, which can lower the model performance.

Federated Learning (FL) is a cross-distributed data modelling method proposed by~\cite{mcmahan2016communication,mcmahan2017communication}. It can establish a global model without exchanging original data among parties. Due to the exponential growth of participated data, the model naturally performs better global robustness and superiority over individual modelling. 

Subsequently,~\cite{cheng2019secureboost} proposed the concept of vertical FL to update it suitable for more realistic scenarios. Since then, many scholars have started to study the application of real FL scenarios and proposed some new algorithms and frameworks, such as SplitNN~\cite{vepakomma2018split}.

~\cite{huang2019patient} reveals the problem of multi-distribution between different data islands through joint clustering and FL. Through five model structure experiments on four different data-sets,~\cite{mcmahan2016communication} demonstrated that the iterative average model can be robust under both IID and non-IID data distribution patterns. However, the iterative approach is not as perfect as imagined. On non-IID data, it requires more rounds to iterate to sufficient convergence, and the final model performance trained with the same optimal parameters always slightly inferior to that obtained under IID distribution.

Almost all FL optimization algorithms are aimed at training a global model. However, in the real scenario, there exist clients who want to train a personalized model by absorbing useful information from others with similar data property. In addition, there are some dishonest participants trying to cheat with useless data to gain a high-qualified model.

Motivated by these real demands, we design a performance-based optimization algorithm, \texttt{FedSmart}, which is automatically updated. Our main contributions are as follows:

1.	Demonstrate the impact and performance of using non-IID data on both FL frameworks and local training.

2.	Adopt independent validation sets in each side instead of shared data sets to improve the model performance on non-IID data.

3.	Propose a new parameter joint method \texttt{FedSmart} to make the multi-party joint value of the stochastic gradient descent close to the unbiased estimate of the complete gradient.

\section{Related Work}

In some cases, due to the advanced nature of some existing machine learning algorithm, the training results based on the non-IID data are still good. However, for some application scenarios, training with non-IID data will have unexpected negative effects based on existing frameworks, such as low model accuracy and convergence efficiency. Because the data on each device is generated independently by the device/user itself, the heterogeneous data of different devices/users have different distribution characteristics and the training data learned by each device during local learning are non-IID. Therefore, how to improve the learning efficiency of non-IID data is of great significance for FL.

\subsection{Average-Based Optimization Algorithm}
To improve the performance of FL and reduce the communication cost~\cite{mcmahan2016communication}, a deep network algorithm \texttt{FederatedAveraging} (\texttt{FedAvg}) based on iterative model average is proposed for non-IID FL, which can be applied to real scenarios. Theoretical analysis and experimental results show that \texttt{FedAvg} is robust to unbalanced and non-IID data, and it also has a low communication cost. Compared with baseline algorithm \texttt{FedSGD}, \texttt{FedAvg} has better practicability and effectiveness. ~\cite{li2020convergence} theoretically clarifies the convergence of \texttt{FedAvg} on non-IID data. Furthermore, \texttt{FedMA} is aimed at settling the heterogeneity problem ~\cite{wang2020federated}.

\subsection{Performance-Based Optimization Algorithm}

The proposal of \texttt{FedAvg} method has a great inspiration for the follow-up researches~\cite{yang2019federated}. ~\cite {zhao2018federated} proposes a data-sharing FL strategy to improve the training of non-IID Data by creating a small portion of the data globally shared between all client devices on a central server. 

Local client computational complexity, communication cost, and test accuracy are three important issues addressed by~\cite{huang2020loadaboost}. It proposes a loss-based AdaBoost federated machine learning algorithm (\texttt{LoAdaBoost}), which further optimizes the local model with high cross-entropy loss before averaging the gradients on the central server.

\texttt{FedProx}, proposed by~\cite{sahu2018federated}, lowers the potential damage to the model caused by non-IID data. It adds a near-end item to optimize the local iteration times. Similarly, \texttt{SCAFFOLD} introduces a new variable combined with gradients, decreasing the variance of local iteration~\cite{karimireddy2019scaffold}.

\section{Approach}
In FL researches, the scholars usually focus on the algorithm framework or the improvement of the global model accuracy. However, we generally do not know the data distribution or data quality of other participants, the heterogeneous data may result in worse performance when added to the global training. 

With these motivations, we propose \texttt{FedSmart}, a new parameter return method. In this mechanism, the FL participant is smart enough to gain information from others who have similar data property. In another aspect, \texttt{FedSmart} can be used to test whether the model from other clients is useful to every client's side. Furthermore, \texttt{FedSmart} can be treated as a kind of latent incentive mechanism, the selfish sides who provide unrealistic or unqualified data will be naturally filtered out via decreasing the weight, only the ones who provide their valuable data can benefit from the group with the similar distributions. 

\subsection{The Information Transfer Framework}

The framework of FL is adopted. There typically exists a \textbf{server}, which controls and publishes the model and jointly deals with the parameters provided by participants. The participants who contribute parameters by doing local model training are called \textbf{clients}.  

\begin{figure}[htbp]
\centerline{\includegraphics[width=1.0\textwidth]{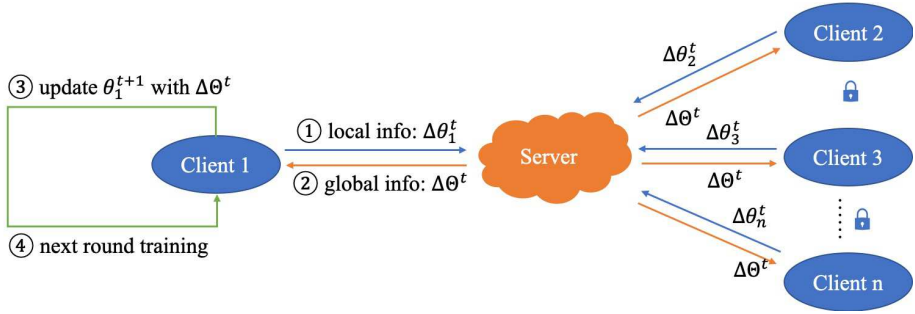}}
\caption{Parameter Update Framework}
\label{Approach}
\end{figure}

All clients do the training respectively using local data. After the model is updated, each client sends the local model information to the server. Clients send the gradient training with their local data to the server; the server packs these changes and sends back, i.e. $\Delta\Theta^t (\Delta\theta_1^t, \Delta\theta_2^t, ..., \Delta\theta_n^t)$ (see Fig.~\ref{Approach}).

\subsection{The Local Model Updating Mechanism}

The local model updating mechanism considers the mutual predicting ability of non-IID data. If all clients train only one global model, it will inevitably lead to distribution or sample size discrimination. \texttt{FedSmart} is designed to update the local model in the form of weights, which makes the model prefer to its self-side data. This approach actually optimizes the server model with the data from each client. 

At the time of initialization, the server initializes the model. When all clients receive the initial model, they will conduct a batch-size training and then launch the information transfer as mentioned above.

\subsection{Performance-Based Weight Allocation}

The weight of the next moment is on the basis of the equation shown below. The performance of all the clients is taken into consideration, the principle, in brief, is that the weight of model will be smartly adjusted to the accuracy of each client.

\begin{equation}
\label{eq1}
||w_i^t||=||w_i^{t-1}+\eta(acc_i^t-acc^t_{median})||_1
\end{equation}

where $acc_i^t$ represents the accuracy of \textit{Client i} on local validation set in round \textit{t} on the validation set, $acc^t_{median}$ is the median of the set of accuracy, and $\eta$ is the learning rate. The weight in round $t$ is allocated according to the weight in the previous round and the change of accuracy in this round. The validation set is extracted from each client with a proportion of $\alpha \in [0,1]$, and only serves for this client.

In \texttt{FedSmart}, we update the model according to the performance on validation set, which makes the model adaptive to self-side data. To conclude, \texttt{FedSmart} actually optimizes model of each client with valuable data from others. 

\begin{algorithm}[htbp]
\caption{\texttt{FederatedSmart} (\texttt{FedSmart})}
\label{alg:FedSmart}
\textbf{Input}: $\theta_i^{t}$: $i$-th client's model parameters at time step $t$;\\
\hspace*{3em}$\Delta \theta_i^{t}$: $i$-th client's model updates at time step $t$\\
%\textbf{Parameter}: Optional list of parameters\\
\textbf{Output}: $\theta_i^{t+1}$: $i$-th client's model parameters at time step $t+1$

\begin{algorithmic}[1] %[1] enables line numbers
\FOR{$i=1$ to $n$}
\STATE aggregate updates for validation: $\theta_i^{t} = \theta_i^{t} + aggregate(\Delta \theta_1^{t}, \Delta \theta_2^{t},...,\Delta \theta_n^{t})$
\STATE{compute model validation accuracy: $acc_i^{t}=evaluate(\theta_i^{t})$}
\ENDFOR

\FOR{$i=1$ to $n$}
\STATE obtain performance-based weight:  $w_i^{t}=weight(acc_1^{t}, acc_2^{t}, ..., acc_n^{t})$
\STATE calculate model parameter update: $\Delta \theta_i^{t}=update(acc_1^{t}, acc_2^{t}, ..., acc_n^{t})$
\ENDFOR

\FOR{$i=1$ to $n$}
\STATE output new model parameters: $\theta_i^{t+1} = \theta_i^{t} + \sum_{i=1}^{n}w_i^{t} \cdot  \Delta \theta_i^{t}$
\ENDFOR

\end{algorithmic}
\end{algorithm}

\section{Experiment}

The experiment settings will be described step by step, including how to deal with the dataset and the experimental settings of \texttt{FedSmart}. Also, we will explain the impact of different parameters on the model performance and demonstrate the mechanism of using validation set.

\subsection{Implementation Details}

The data that concerned with the performance evaluation is the simulated datasets of MIMIC-III database \cite{pollard2016mimic,johnson2016mimic}, which contains the health information for critical care patients at a large tertiary care hospital in the U.S. The data cleansing process is following~\cite{huang2020loadaboost}. 

The experimental data structure is shown in Table~\ref{tab:statistics}.

\begin{table*}[htbp]
\begin{threeparttable}
\caption{Summary of Experiment Dataset}
\label{tab:statistics}
\newcommand{\tabincell}[2]{\begin{tabular}{@{}#1@{}}#2\end{tabular}}
\centering
\setlength{\tabcolsep}{3mm}
    \begin{tabular}{lcr}  
        \toprule
    Feature	& Representation & Count\\
        \midrule
    $SUBJECTID$$^{\rm a}$ & IDs ranging from 2 to 99,999 & 21000 selected from 38962\\
        \midrule
    $GENDER$$^{\rm b}$ &\tabincell{c} {0: female  1: male} & 9900/12000\\
        \midrule
    $AGE$$^{\rm b}$ & \tabincell{c} {0: age less than or equal to 65\\1: age greater than 65} & 9903/11997\\
        \midrule
    $MORTALITY$$^{\rm c}$ & \tabincell{c} {0: survive  1: death} & 10785/11115\\
        \midrule
    $DRUGS$$^{\rm d}$ & \tabincell{c} {0: not prescribed to patients\\1: prescribed} & 8 dimensions\\
        \bottomrule
    \end{tabular}

    \begin{tablenotes}

    \item$^{\rm a}$\textit{SUBJECTID} is the primary key. \item$^{\rm b}$\textit{GENDER} and \textit{AGE} indicate basic information about the patients.
    \item$^{\rm c}$\textit{MORTALITY} indicates survival status. The original distribution of \textit{MORTALITY} is biased, it is three-times up-sampled. \item$^{\rm d}$\textit{DRUGS} represents each patient’s usage of the particular drugs during the first 48 hours in the ICU.
  
    \end{tablenotes}

\end{threeparttable}

\end{table*}

\subsection{Experiment Settings}

 To illustrate the limited performance of FL on non-IID data, the data are constructed as a collective form of six heterogeneous data sets. In detail, \textit{Client1} and \textit{Client4}, \textit{Client2} and \textit{Client5}, \textit{Client3} and \textit{Client6} in pairs share a similar data distribution respectively.
 
 The validation set proportion $\alpha$ is set to 0.25 in default all through the experiment.

\subsection{Results}

The essence of centralized training is to aggregate the data of all parties together to improve the accuracy of the model by increasing the amount of data, so the results of centralized training are often higher than the results of each client training on their datasets alone. However, when the data are non-IID, centralized training will be hard to balance the results. The model tends to favor the groups with large samples or with simple distributions, so the established global model is undoubtedly unfair to other groups.

\subsubsection{FedSmart v.s. Local Training}

\begin{figure} 
\begin{minipage}[t]{0.5\linewidth} 

\centering 
\includegraphics[width=2.0in]{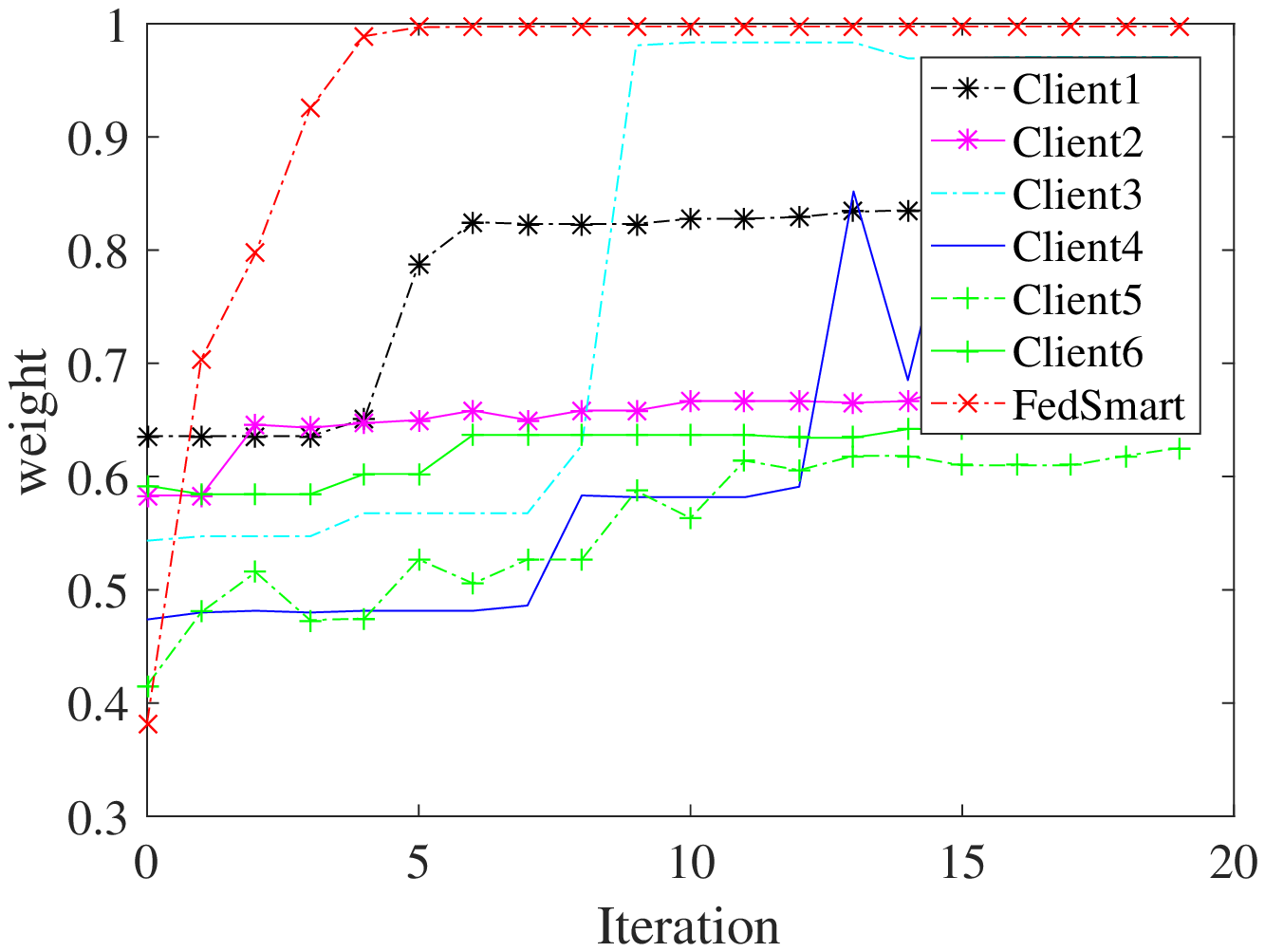} 
\caption{\texttt{FedSmart} v.s. Local Training} 
\label{GlobalLocal} 
\end{minipage} 
\begin{minipage}[t]{0.5\linewidth} 

\centering 
\includegraphics[width=2.25in]{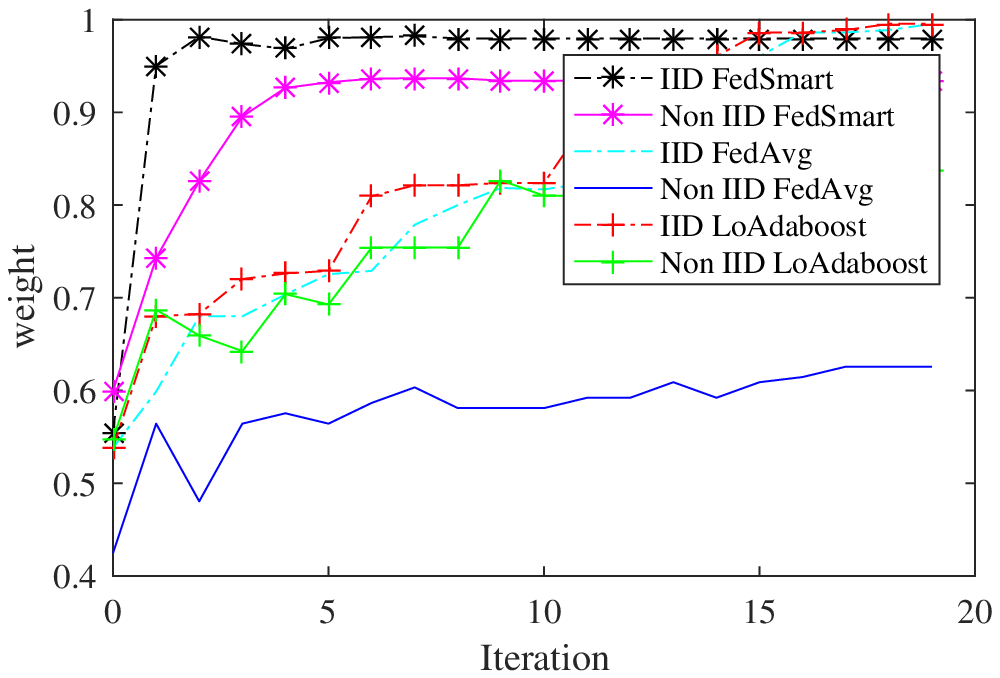} 
\caption{\texttt{FedSmart} v.s. \texttt{FedAvg} v.s. \texttt{LoAdaBoost}} 
\label{compareAlgorithm} 
\end{minipage} 
\end{figure}

The model trained with \texttt{FedSmart} outperforms the local six ones (see Fig.~\ref{GlobalLocal}), which is in the expectation that all FL participants will gain a better model within the information sharing framework than only using their own data. Because compared to the individual, working in a team, sharing the information of data, i.e. in the framework of FL, everyone tends to gain something as a contributor. 

\subsubsection{FedSmart v.s. FedAvg v.s. LoAdaBoost}

To illustrate the effectiveness of \texttt{FedSmart}, we will do a comparison among \texttt{FedSmart}, \texttt{FedAvg} and \texttt{LoAdaBoost}. In \texttt{FedAvg}, the server only receives the model parameters and returns the updated model parameters, and there is no interactive updating mechanism in \texttt{FedAvg}. \texttt{LoAdaBoost} receives the loss and parameters of the model, and combines the information of the two to update the weight of the previous iteration~\cite{huang2020loadaboost}. FedSmart adopts different parameter combinations to update the model to make it approximate to the unbiased estimate of the complete gradient. The result is shown in Fig.~\ref{compareAlgorithm}.

It can be seen that no matter what FL optimization algorithm, the performance on IID data always outperforms non-IID ones. One of the most important incentives of FL optimization algorithm is to decrease the influence of data distribution, i.e. the performance reduction on non-IID data. Also, \texttt{FedSmart} uses the accuracy of the validation set to measure the similarity of the distribution, establishes multiple models by adjusting the weights of different client models, and establishes multiple models on multiple clients only through the encrypted parameter exchange. The result shows that model performance is significantly better than \texttt{FedAvg}, and moderately better than \texttt{LoAdaBoost}.

\subsubsection{FedSmart}
\begin{figure*}[htbp]
\centering
\begin{minipage}[b]{1\textwidth}
    \subfigure[\textit{Client1}]{
        \includegraphics[width=0.31\textwidth]{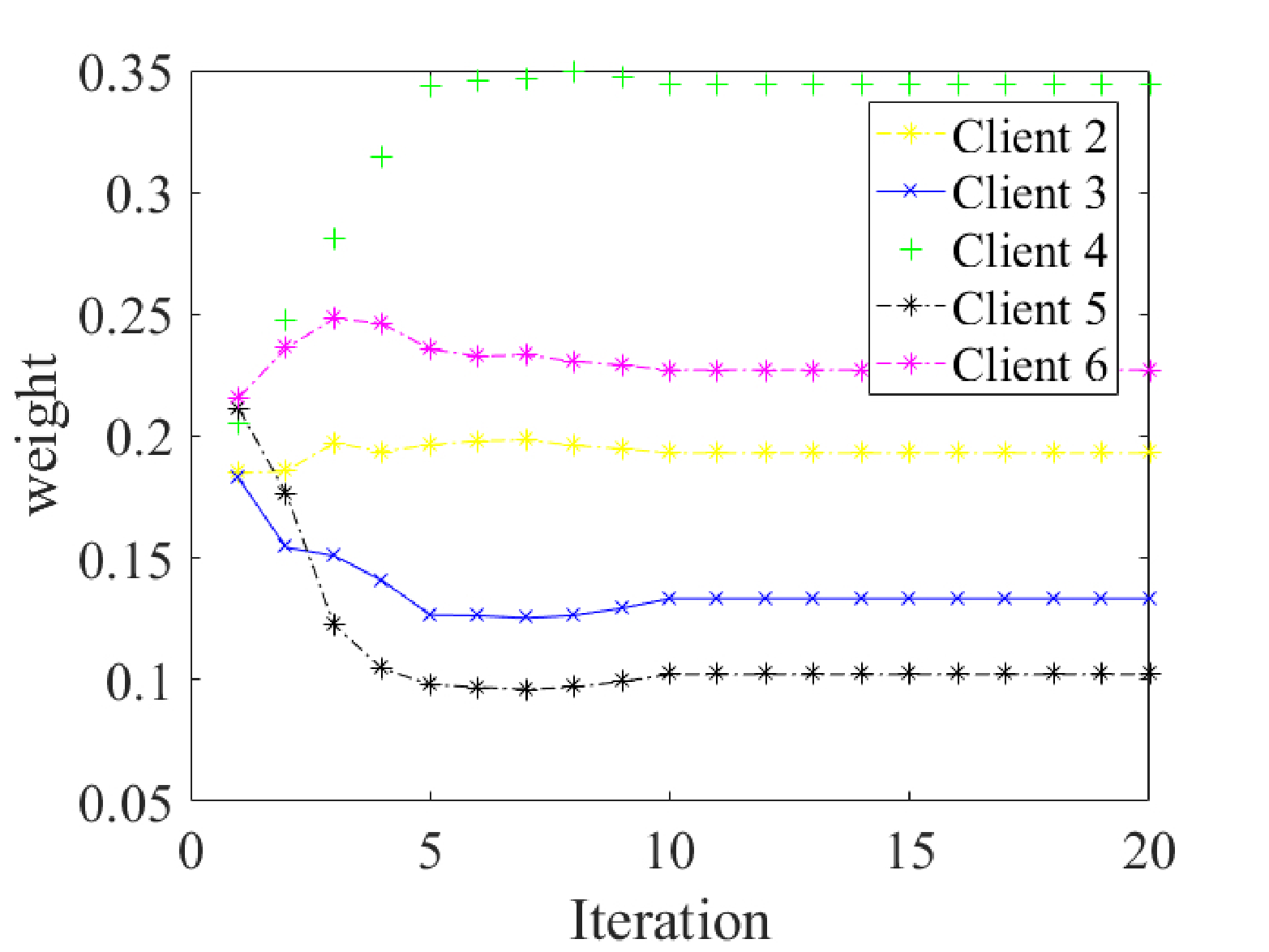}
    }
    \subfigure[\textit{Client2}]{
        \includegraphics[width=0.31\textwidth]{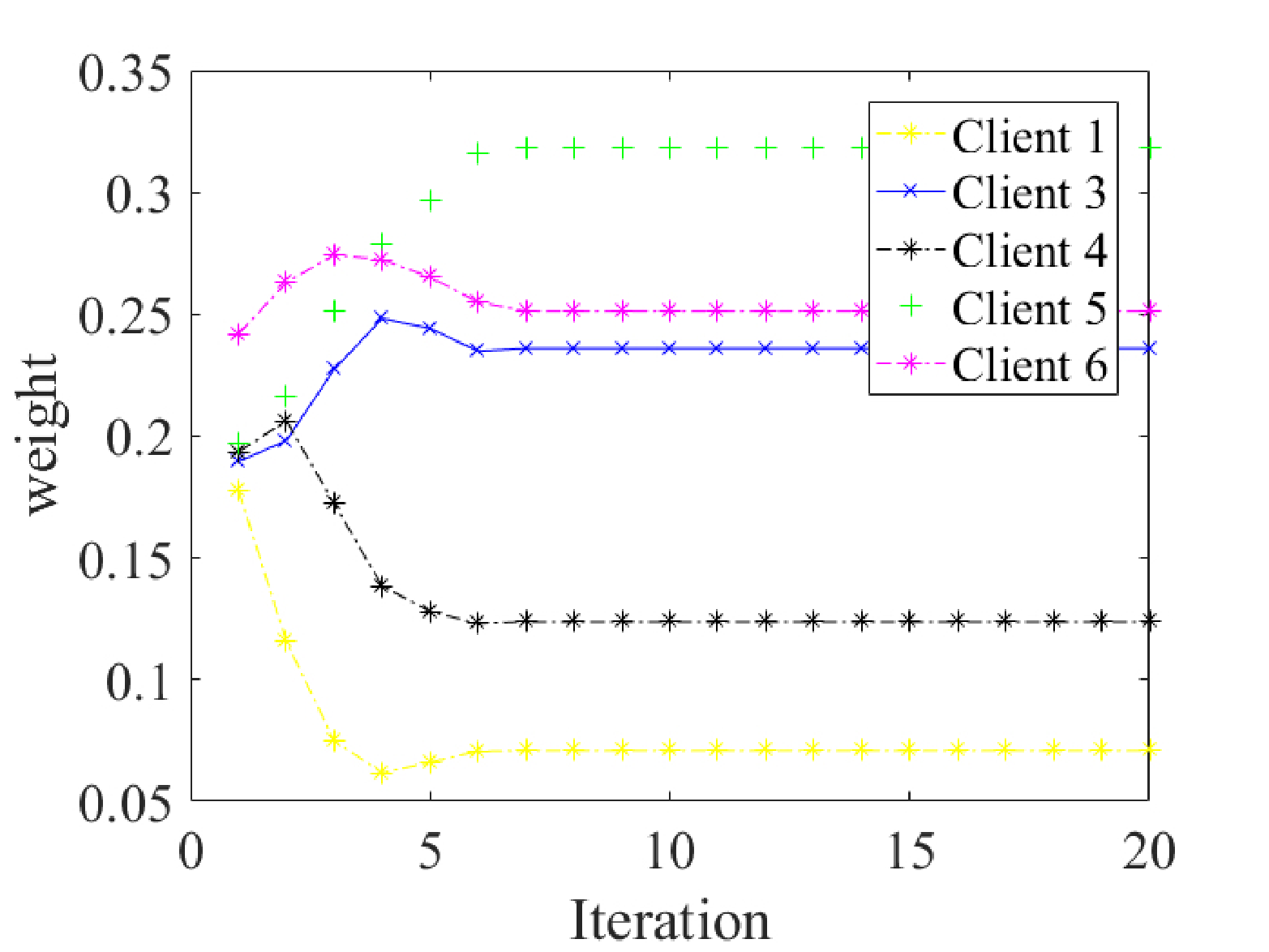}
    }
    \subfigure[\textit{Client3}]{
        \includegraphics[width=0.31\textwidth]{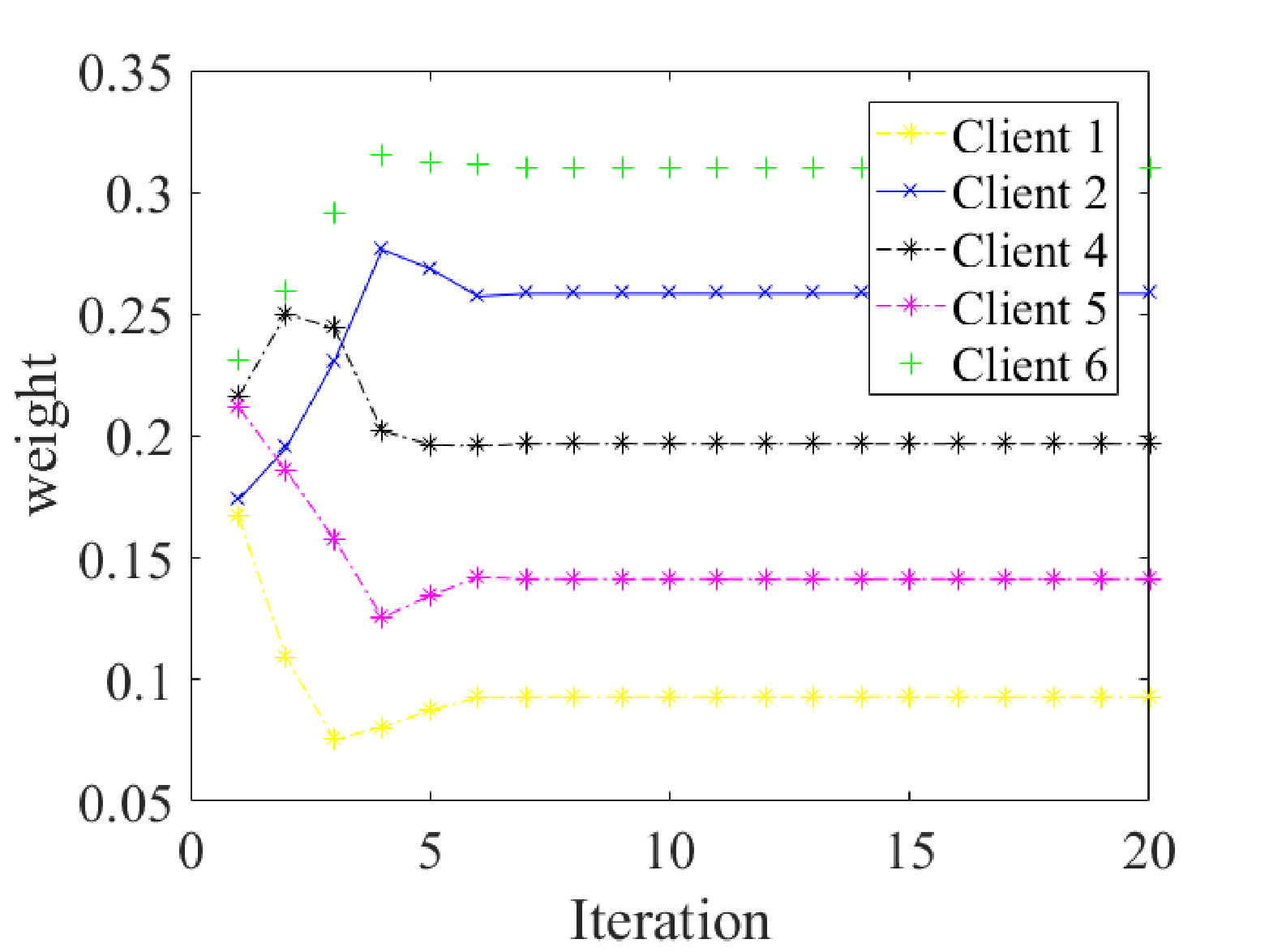}
    }
\end{minipage}

\begin{minipage}[b]{1\textwidth}
    \subfigure[\textit{Client4}]{
        \includegraphics[width=0.31\textwidth]{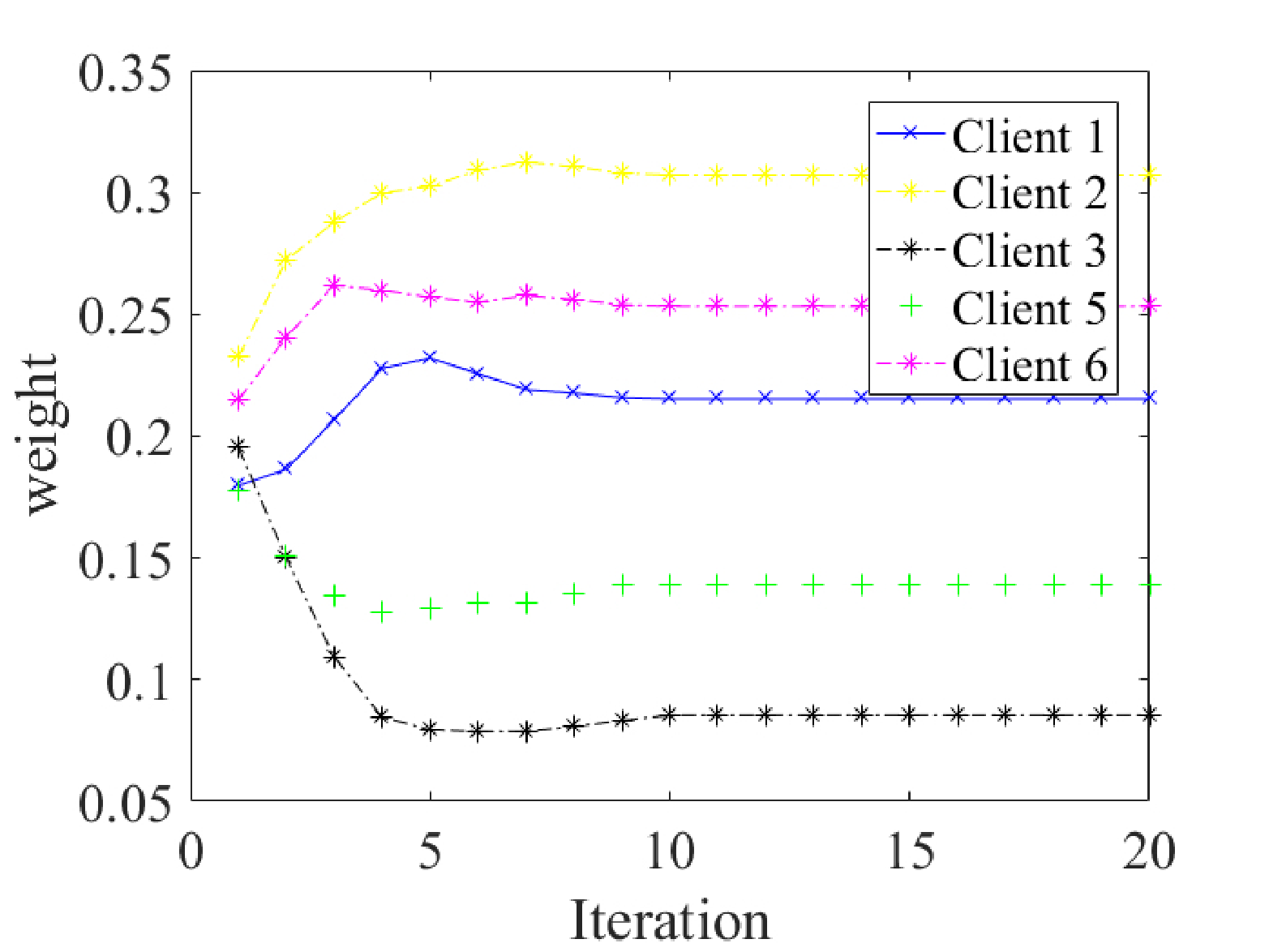}
    }
    \subfigure[\textit{Client5}]{
        \includegraphics[width=0.31\textwidth]{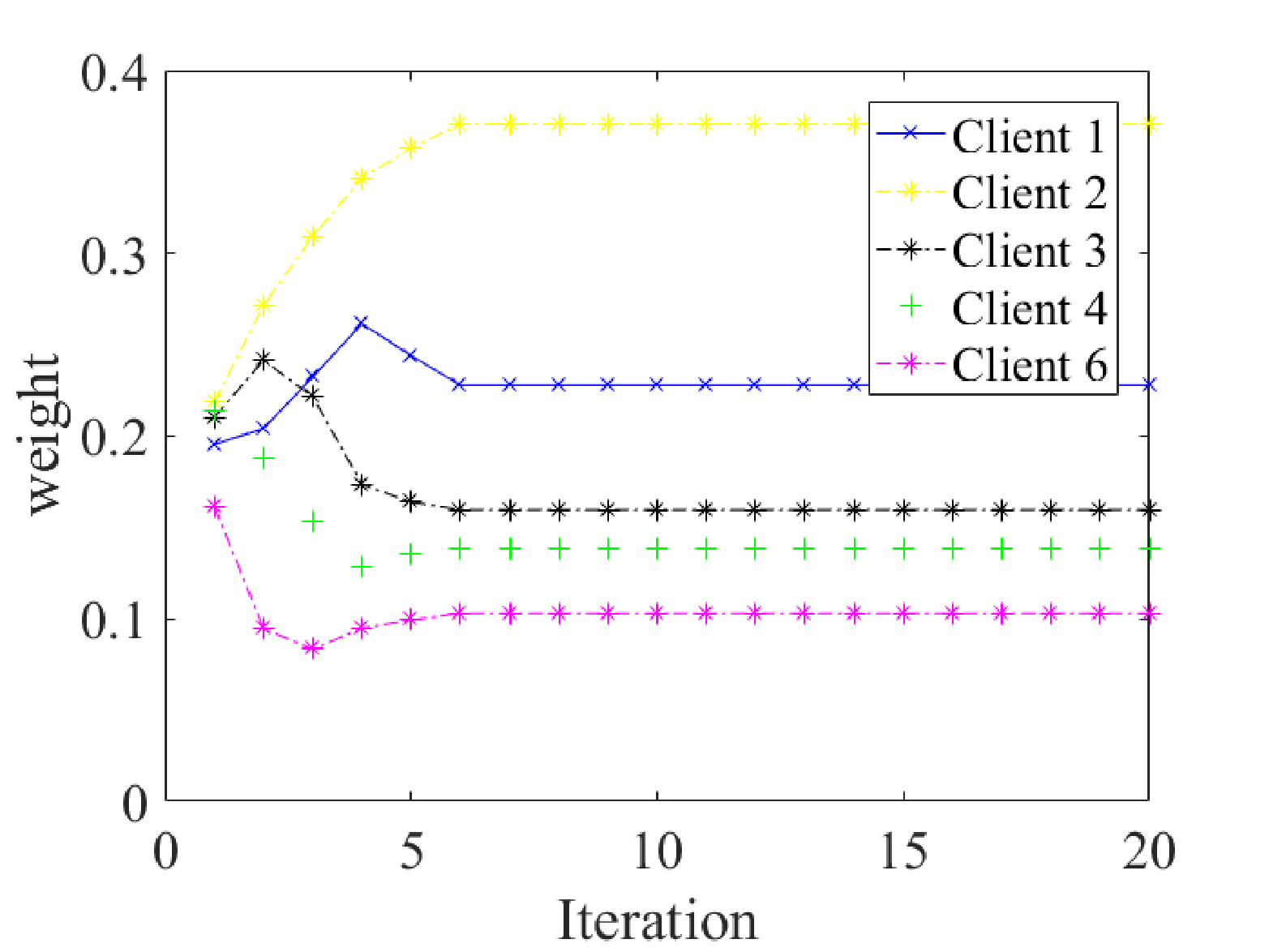}
    }
    \subfigure[\textit{Client6}]{
        \includegraphics[width=0.31\textwidth]{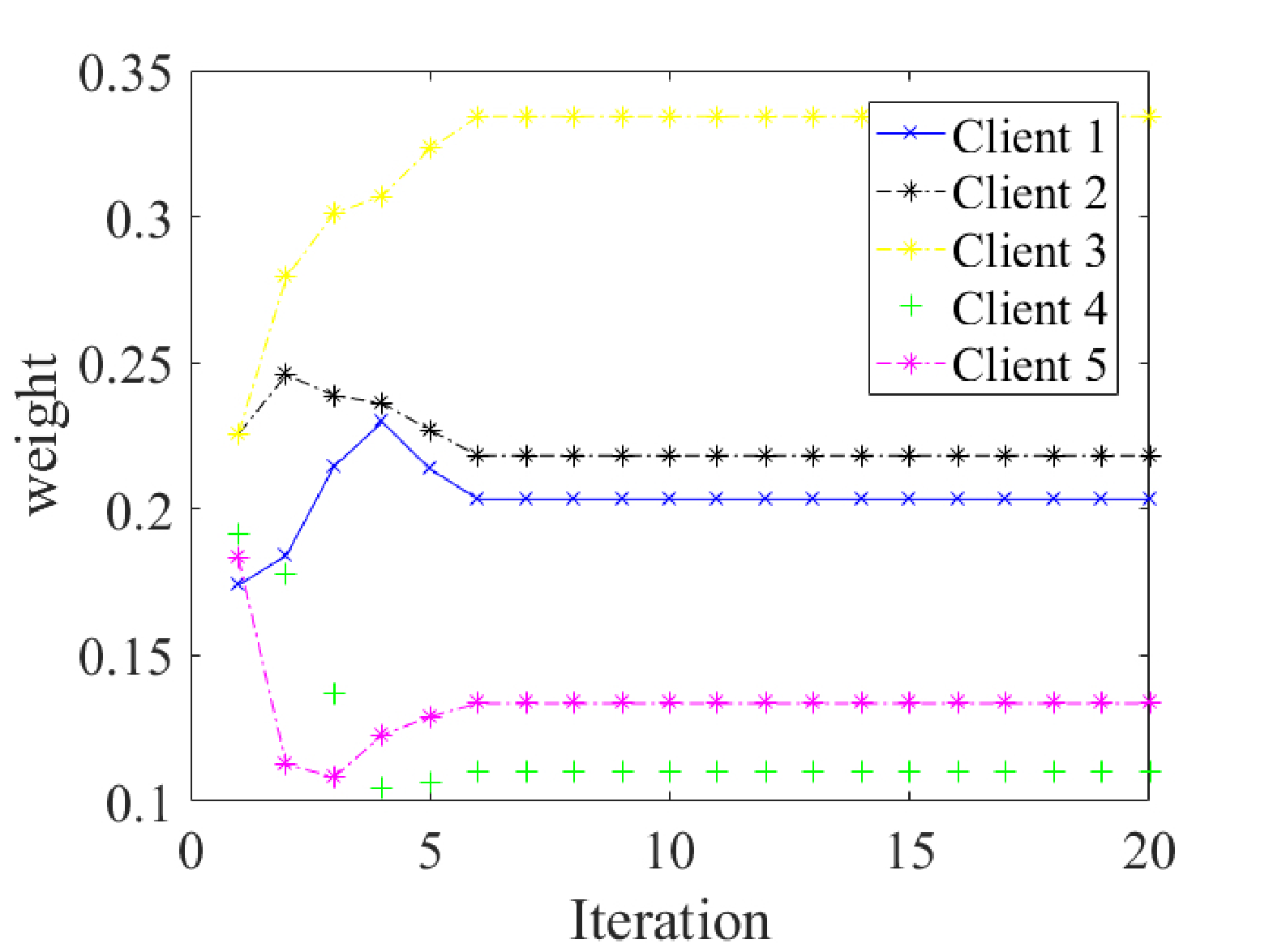}
    }
\end{minipage}
\caption{The Process of Weight Allocation. The weight appears to change in pairs: \textit{Client1} and \textit{Client4}, \textit{Client2} and \textit{Client5}, \textit{Client3} and \textit{Client6}.}
\label{WAC-all}
\end{figure*}

\texttt{FedSmart} considers one party’s distribution without repeatedly making compromises on multiple distributions. To further explain the working mechanism and performance of \texttt{FedSmart}, the process of the parameter joint weight changing during the training process is shown as in Fig.~\ref{WAC-all}. The weight appears to change in pairs: \textit{Client1} and \textit{Client4}, \textit{Client2} and \textit{Client5}, \textit{Client3} and \textit{Client6}, which is in accordance with our experimental settings, indicating that \texttt{FedSmart} is figuring out good data.

We can observe that in the FL, there still exists the unbalanced performance improvement on some sides due to the difference in distributions. Because normally we only have one global model to be established, to reduce the global loss and improve the accuracy, there is inevitably a decrease in the performance improvement caused by the fact that one of the distributions is ignored to some extent. As long as there is only one global model, attend to one thing and lose sight of another must occur. Therefore, for the non-IID data, it is necessary to consider how to create multiple models suitable for different distributions, and then make FL more universal.

\section{Conclusion}

Federated Learning is raising attention in both academics and industry, as it is a way to solve the isolated island problem and a solution to privacy-preserving. We propose a performance-based parameter return method \texttt{FedSmart}. It is different from the general idea that FL shares one global model. Instead, \texttt{FedSmart} establishes multiple models by treating each client as a server to make its own model perform the best. We use the simulated MIMIC-III data and separate it into six non-IID data-sets to do the FL. The experimental result shows that \texttt{FedSmart} can have better performance than \texttt{FedAvg} and even centralized training method. \texttt{FedSmart} can be extended to the industries’ data training scenarios. 

In the continuation of our study, to compensate for this shortcoming and minimize the leakage of privacy caused by model delivery, \texttt{FedSmart} can use the drop-out-like mechanism to make it difficult for training participants to obtain effective information from the changes of the model. Also, we will improve and explore the \texttt{FedSmart} algorithm to make it to be generally stable and adaptable for both IID and Non-IID datasets, to tackle the root of problems for FL frameworks.

\subsubsection{Acknowledgements.} This paper is supported by National Key Research and Development Program of China under grant No.2018YFB1003500, \\No.2018YFB0204400 and No.2017YFB1401202. The corresponding author is Jianzong Wang, jzwang@188.com.

\end{document}